\documentclass[sigconf, prologue,table]{acmart} 

\usepackage{amsmath}
\usepackage{tikz}
\usepackage{algorithm}
\usepackage{algpseudocode}
\usepackage{float} 
\usepackage{booktabs}
\usepackage{tabularx}
\usepackage{array}
\usepackage{graphicx}
\usepackage{subcaption}
\usepackage{multirow}
\usepackage{enumitem}
\usepackage{array}
\usepackage{caption}
\usepackage{balance}
\usepackage{placeins} 

\newcommand{\highlightbox}[1]{%
  \tikz[baseline=(char.base)] \node[draw=red, thick, inner ysep=0.7pt, inner xsep=2pt] (char) {#1};%
}
\AtBeginDocument{%
  }

\copyrightyear{2025}
\acmYear{2025}
\setcopyright{cc}
\setcctype{by-nc}
\acmConference[CIKM '25]{Proceedings of the 34th ACM International
Conference on Information and Knowledge Management}{November 10--14,
2025}{Seoul, Republic of Korea}
\acmBooktitle{Proceedings of the 34th ACM International Conference on
Information and Knowledge Management (CIKM '25), November 10--14, 2025,
Seoul, Republic of Korea}
\acmISBN{979-8-4007-2040-6/2025/11}
\acmDOI{10.1145/3746252.3761250}

\begin{document}
\title{From Patterns to Predictions: A Shapelet-Based Framework for Directional Forecasting in Noisy Financial Markets}

\author{Juwon Kim}
\affiliation{%
  \institution{Pohang University\\
of Science and Technology}
  \city{Pohang}
  \country{Republic of Korea}}
\email{juwona@postech.ac.kr}

\author{Hyunwook Lee}
\affiliation{%
  \institution{Pohang University\\
of Science and Technology}
  \city{Pohang}
  \country{Republic of Korea}}
\email{gusdnr0916@unist.ac.kr}

\author{Hyotaek Jeon}
\affiliation{%
  \institution{Pohang University\\
of Science and Technology}
  \city{Pohang}
  \country{Republic of Korea}}
\email{taek98@postech.ac.kr}

\author{Seungmin Jin}
\affiliation{%
  \institution{Ulsan National Institute\\ of Science and Technology}
  \city{Ulsan}
  \country{Republic of Korea}}
\email{skyjin@unist.ac.kr}

\author{Sungahn Ko}
\affiliation{%
  \institution{Pohang University\\
of Science and Technology}
  \city{Pohang}
  \country{Republic of Korea}}
\email{	sungahn@postech.ac.kr}
\authornote{Corresponding author.}

\begin{abstract}
Directional forecasting in financial markets requires both accuracy and interpretability. Before the advent of deep learning, interpretable approaches based on human-defined patterns were prevalent, but their structural vagueness and scale ambiguity hindered generalization. In contrast, deep learning models can effectively capture complex dynamics, yet often offer limited transparency. To bridge this gap, we propose a two-stage framework that integrates unsupervised pattern extracion with interpretable forecasting. (i) \textbf{SIMPC} segments and clusters multivariate time series, extracting recurrent patterns that are invariant to amplitude scaling and temporal distortion, even under varying window sizes. (ii) \textbf{JISC-Net} is a shapelet-based classifier that uses the initial part of extracted patterns as input and forecasts subsequent partial sequences for short-term directional movement. Experiments on Bitcoin and three S\&P 500 equities demonstrate that our method ranks first or second in \textbf{11 out of 12 metric--dataset combinations}, consistently outperforming baselines. Unlike conventional deep learning models that output buy-or-sell signals without interpretable justification, our approach enables transparent decision-making by revealing the underlying pattern structures that drive predictive outcomes.

\end{abstract}


\maketitle

\section{Introduction}

Accurately forecasting market movements such as stock prices with deep learning is difficult due to the frequent volatility and low signal-to-noise ratio inherent in financial markets. For instance, Rossi \cite{Rossi2013} notes that exchange rate forecasts often fail to outperform a simple random walk. Even advanced models often fail to outperform naive baselines using only the last observed price \cite{zeng2022dlinear}. These challenges have led to numerous stock prediction approaches, broadly classified into two types: (i) internal time series-based models, which use only historical prices and technical indicators \cite{fan2024stockmixer,qin2017alstm}; and (ii) hybrid models, which combine internal price dynamics with external signals such as financial news, social media, or macroeconomic indicators \cite{li2023pen,choudhary2024predictive}.
In this study, following the efficient market hypothesis that asset prices reflect all available information, we focus solely on price data to capture the intrinsic dynamics of stock price movements. 
Given the universal availability of price data and independence from exogenous inputs, this approach is applicable across diverse stocks and markets. However, price-only modeling faces three major challenges. First, financial time series are highly noisy with weak long-term signals, making it difficult to separate meaningful patterns from random fluctuations. Second, recurrent patterns appear at varying scales and durations. We term these variations \emph{\textbf{amplitude-scaling}}, denoting proportional changes in both magnitude and duration, and \emph{\textbf{temporal-distortion}}, referring to non-uniform changes in the timing or speed of specific segments within a pattern. These two types encompass the canonical distortions explored in prior work~\cite{lee2022tilde, esling2012time}, offering a structured perspective for interpreting recurrent patterns within financial data. Third, deep neural networks can model complex market dynamics but offer limited \textit{\textbf{interpretability}} due to their black-box nature.

To overcome these limitations, we introduce a two-stage framework that extracts and leverages interpretable patterns from noisy \textit{\textbf{multivariate}} financial time series. In the \textbf{pattern extraction} stage, we propose the \emph{Selective Invariant Multivariate Pattern Clustering} (SIMPC) module, which segments and clusters multivariate subsequences using \emph{Dynamic Time Warping} (DTW)~\cite{berndt1994using}. SIMPC identifies patterns invariant to amplitude scaling and temporal distortion, even under dynamic window conditions. By incorporating multivariate inputs, it enables fine-grained contextual differentiation. In the \textbf{pattern detection} stage, we propose the \emph{Joint-variable Invariant Shapelet-Classification Network} (JISC-Net), which combines DTW with a \textit{Multi-length-input dilated causal Convolutional Neural Network} (Mdc-CNN) encoder~\cite{Li2021ShapeNet} to learn representations robust to temporal distortions. JISC-Net identifies and filters meaningful pattern instances through a two-stage filtering process, excluding noisy or non-pattern segments. 
Empirical evaluations on four financial assets show that our method significantly improves directional accuracy and trading performance over competitive baselines. By focusing on localized subsequences that align with empirically extracted pattern structures, rather than modeling the entire time series, the proposed framework enhances both interpretability and practical utility in real-world financial forecasting.

 
\section{Related Work}
\paragraph*{\textbf{Time Series Forecasting Models.}}
Relevant forecasting models include multivariate approaches and stock-specific deep learning methods. \textbf{Multivariate time series forecasting} has made substantial progress in modeling complex temporal and cross-variable dependencies. FEDformer~\cite{zhou2022fedformer} extends Transformer-based forecasting by leveraging seasonal-trend decomposition in the frequency domain, while PatchTST~\cite{Nie2023PatchTST} divides time series into patch tokens and applies channel-independent self-attention to capture long-term dependencies. In addition, TimesNet~\cite{wu2023timesnet} employs temporal 2D variation blocks to capture multiscale patterns in the time-frequency space. 
Despite their effectiveness on standard benchmarks, these models often underperform on noisy financial time series due to their reliance on dense temporal encoding, which increases the risk of overfitting.
In response, \textbf{stock-specific architectures} have been proposed to address these domain challenges. 
The \textit{State-Frequency Memory} (SFM) network~\cite{zhang2017stock} decomposes temporal data into frequency components to model stock price dynamics but struggles with non-stationary conditions. StockMixer~\cite{fan2024stockmixer} utilizes a compact MLP for patch-based mixing of multivariate inputs, capturing local patterns.  MASTER~\cite{li2024master} integrates a market-guided gating mechanism within a Transformer framework, dynamically selecting features and aggregating information across stocks to capture market-driven relationships. More recently, SAMBA~\cite{mehrabian2025mamba} combines continuous-state sequence modeling with a dynamic stock graph for scalable long-term forecasting. Even with these advancements, similar to time series models , stock-specific approaches remain black boxes with limited transparency, rarely utilizing human-interpretable patterns for predictive decisions.

\paragraph*{\textbf{Pattern Clustering in Time Series.}}
Unsupervised and weakly supervised methods have been developed to identify recurring temporal structures. For example, \textit{Deep Temporal Contrastive Clustering} (DTCC)~\cite{DTCC} introduces contrastive learning into time series clustering for the first time, combining instance-level and cluster-level contrastiveness within a self-supervised autoencoder framework. By jointly optimizing reconstruction, clustering, and contrastive losses, DTCC improves representation quality for general time series clustering tasks. However, it does not specifically address the challenges posed by financial time series, such as amplitude scaling and temporal distortion.
A particularly relevant precursor to our approach is the \textit{Scale-Invariant Subsequence Clustering} (SISC) algorithm~\cite{Huang2024FTSDiffusion}, which partitions financial time series into variable-length segments and clusters them into shape-based groups that are invariant to both amplitude and duration. While effective at identifying latent motifs, SISC is limited to univariate inputs and may admit noisy or irrelevant segments.
\textbf{Our proposed SIMPC} extends this line of work by accommodating multivariate inputs, initializing clusters with domain-informed prototypes, and explicitly filtering out low-information subsequences.

\paragraph*{\textbf{Shapelet-Based Classification.}}
Shapelet-based methods provide interpretable representations for time series classification by identifying discriminative subsequences. The concept was first introduced by Ye and Keogh~\cite{ye2009} for univariate data, and later extended to learnable shapelets~\cite{Grabocka2014} and binary representations for multiclass classification~\cite{Bostrom2017Binary}. ShapeNet~\cite{Li2021ShapeNet} further generalized these methods to multivariate series by incorporating dilated causal convolutions and triplet loss to learn embeddings of variable-length shapelets. Building on this foundation, \textbf{our proposed JISC-Net} incorporates a DTW-based triplet loss to enhance robustness against temporal distortions common in financial data, and treats multivariate observations as unified tokens to maintain joint temporal dependencies across variables.
\section{Problem Statement}

\begin{figure*}[t]
    \centering
    \includegraphics[width=1.0\linewidth]{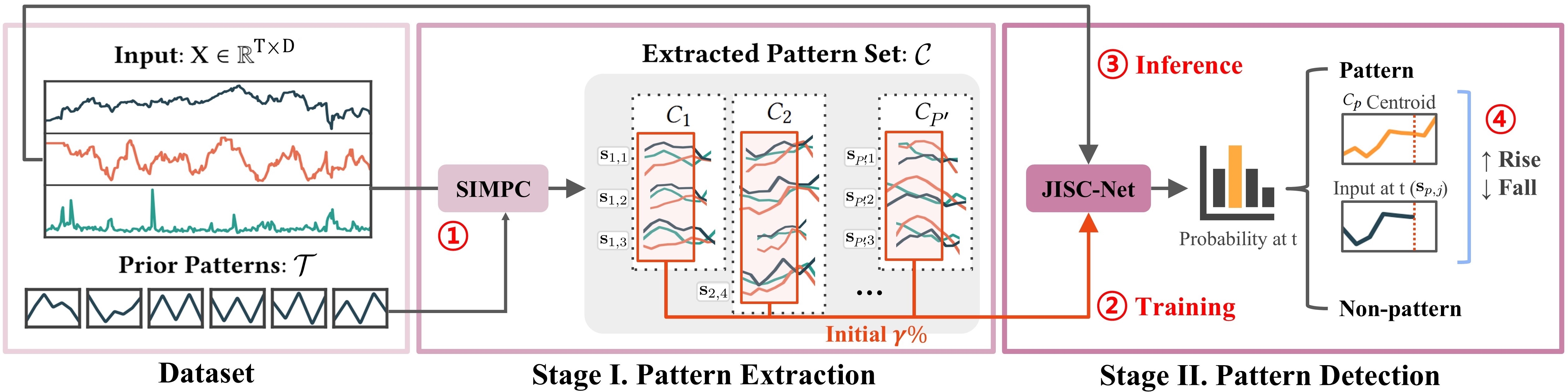} 
    \caption{Overview of the proposed two-stage framework. Stage 1: \textbf{SIMPC} extracts robust, recurring patterns from multivariate time series under amplitude scaling and temporal distortion. Stage 2: \textbf{JISC-Net} performs pattern classification using the initial portion of each pattern subsequence and predicts the directional movement of its remaining part.}
    \Description{A diagram illustrating a two-stage time series prediction framework with pattern extraction and classification modules.}
    \label{fig:your_label}
\end{figure*}

The goal is to identify and detect recurring patterns embedded in noisy \textbf{multivariate} time series that exhibit \textbf{amplitude-scaling} and \textbf{temporal-distortion} characteristics, in a fully \textbf{unsupervised} manner, and to predict future directional movement (rise or fall) based on these patterns. This is accomplished through two stages: pattern extraction and classification, as illustrated in Figure~\ref{fig:your_label}.

\textbf{Pattern Extraction.} 
Let $\mathbf{X} = [\mathbf{x}_1, \ldots, \mathbf{x}_T] \in \mathbb{R}^{T \times D}$ be a multivariate time series, where $\mathbf{x}_t \in \mathbb{R}^D$ denotes the feature vector at time $t$. During pattern extraction, SIMPC segments $\mathbf{X}$ and derives $P$ pattern clusters, denoted as
\(\mathcal{C} = \{ C_1, C_2, \ldots, C_P \}\).
Each cluster $C_p$ comprises multiple temporally contiguous subsequences associated with pattern $p$:
\[
C_p = \{ \mathbf{s}_{p,1}, \mathbf{s}_{p,2}, \ldots\}, \quad \mathbf{s}_{p,j} \in \mathbb{R}^{\ell \times D}
\]
where $\mathbf{s}_{p,j}$ denotes the $j$th subsequence in cluster $C_p$, with variable length $\ell \in [L_{\min}, L_{\max}]$.  
Cluster sizes may vary depending on the frequency of pattern occurrences in the time series.
Each subsequence $\mathbf{s}_{p,j}$ is a contiguous segment of $\mathbf{X}$, starting at a specific time step. To ensure non-overlapping pattern boundaries, we require that the start time of each subsequence be unique across all clusters. 
Since the $P$ clusters may contain redundant or semantically similar patterns, we perform cluster merging in the final stage of SIMPC, which results in a refined pattern set of size~$P'$.

\textbf{Pattern Detection.}
The classifier JISC-Net is trained on the extracted pattern set $\mathcal{C}$ to map an input subsequence $\mathbf{s}_{p,j}$ to a probability distribution over $P'$ pattern types. Importantly, only the initial $\gamma$ portion of each subsequence is used as input to JISC-Net, where \( \gamma \in (0, 1) \) denotes the proportion of the sequence that is utilized during classification. The remaining \((1 - \gamma)\) portion can be reasonably inferred once the pattern is detected, as its structure has already been captured during pattern extraction.
To ensure reliable classification predictions, we apply two filtering mechanisms. During training, label-level filtering with the \textit{Kolmogorov-Smirnov} (K-S) test excludes statistically insignificant subsequences from the trading candidate set. During inference, confidence-based thresholding (top-$x$\%) removes ambiguous or non-pattern instances. As a result, JISC-Net can reliably identify and reject inputs that do not correspond to meaningful patterns.

\section{SIMPC: Pattern Extraction}

\textit{Selective Invariant Multivariate Pattern Clustering} (SIMPC) extracts patterns by segmenting a multivariate time series into variable-length candidates and clustering only the most informative ones. It extends the \textit{Segmentation-Invariant Subsequence Clustering} (SISC) algorithm~\cite{Huang2024FTSDiffusion}, which is a K-means++-style algorithm that greedily segments univariate series into variable-length windows and clusters them using DTW.
However, SISC suffers from key limitations: it is restricted to univariate inputs, initializes the first centroid randomly as in K-means++, and clusters all segments regardless of informativeness.  \textbf{SIMPC addresses these issues by (1)} generalizing DTW to multivariate inputs to capture inter-variable dependencies. This enables the model to distinguish semantically distinct behaviors that may appear identical in a single variable, enhancing interpretability through richer multivariate context; \textbf{(2)} incorporating domain knowledge via canonical chart pattern prototypes for centroid initialization (see $\mathcal{T}$ in Fig.~\ref{fig:your_label}); and \textbf{(3)} pruning low-information segments and discarding sparse clusters to suppress noise.
We proceed by detailing each component of the algorithm.

\paragraph{\textbf{1. Pre-processing by Kernel Regression}}
To suppress short-term volatility, we smooth the multivariate input \(\mathbf{X}\) using Nadaraya-Watson kernel regression~\cite{nadaraya1964estimating} for each variable independently.
\begin{equation}
\label{equ:regression}
\tilde{\mathbf{x}}_t
= \frac{\sum_{s=1}^{T} K_h(t-s)\,\mathbf{x}_s}{\sum_{s=1}^{T} K_h(t-s)},
\qquad
K_h(u) = K\!\left(\frac{u}{h}\right),
\end{equation}
where $K(\cdot)$ is a Gaussian kernel and $h$ is the bandwidth.  
This yields the denoised sequence \(\tilde{\mathbf{X}}\), subsequently used for clustering.

\paragraph{\textbf{2. Domain Adaptation via Canonical Chart Patterns}}

Empirical evidence shows that common behavioral biases recur in similar forms across asset classes, supporting the use of canonical stock rules as a cross-market prior~\cite{lo2004adaptive}. Motivated by this, we incorporate domain knowledge into clustering by constructing the initial centroid set using \textit{traditional chart patterns} (see Appendix~\ref{appendix:chart-patterns}). This step is performed independently of the main clustering pipeline using a separate historical dataset, from which we extract multivariate time series segments that satisfy canonical chart rules~\cite{lo2000foundations}. Prior to extraction, each variable is smoothed with kernel regression (Eq.~\ref{equ:regression}) to mitigate short-term volatility.

We first identify pattern-conforming intervals based on \textit{price} alone, following structural constraints~\cite{lo2004adaptive}, and then collect additional market variables over the same spans. Each extracted segment is independently normalized per variable using Min-Max scaling, since patterns depend on relative shape rather than absolute magnitude. After normalization, all segments are linearly interpolated to a common reference length $L$.
To compute a representative prototype for each chart pattern class, we collect all normalized \textit{price} sequences assigned to that class and aggregate them using \textit{DTW Barycenter Averaging} (DBA). This yields an average price trajectory that preserves structural features while temporally aligning the sequences. The alignment for each sequence is captured in its own warping matrix, which records how that price sequence is matched to the class-specific average. We then multiply its companion variables by each sequence’s warping matrix, thereby temporally synchronizing all dimensions of the multivariate sequence with the average price trajectory.
Finally, for each pattern, we compute the multivariate prototype $\mathcal{T}_i$ by taking the element-wise mean across all aligned multivariate segments.

\paragraph{\textbf{3. Adapted K-means++ Initialization}}

A user-defined number $m$ of prototypes $\{\mathcal{T}_1, \dots, \mathcal{T}_m\}$ is retained as \textbf{fixed seeds}. Since SIMPC is designed to form a total of $P$ pattern clusters, the remaining $P-m$ centroids are initialized using this procedure.
We enumerate all valid start indices $s \in \{1, \dots, T - L_{\max} + 1\}$ in the smoothed input $\tilde{\mathbf{X}}$ to extract candidate segments of length $L_{\max}$:
\[
\mathbf{q}_{s, L_{\max}} = \tilde{\mathbf{X}}_{s : s + L_{\max} } \in \mathbb{R}^{L_{\max} \times D},
\]
Each segment $\mathbf{q}_{s, L_{\max}}$ is Min-Max normalized per variable  to yield $\hat{\mathbf{q}}_{s, L_{\max}}$, emphasizing relative shape changes over absolute values.
We then compute the minimum DTW distance between each normalized candidate $\hat{\mathbf{q}}_{s, L_{\max}}$ and the current centroid set $\mathcal{P}$:
\[
d_s = \min_{C \in \mathcal{P}} d_{\text{DTW}}(\hat{\mathbf{q}}_{s, L_{\max}}, C).
\]
Sampling probabilities are defined as
\( p_s = d_s \bigl/ \sum_{i \in \mathcal{S}} d_i\), where $\mathcal{S}$ denotes the set of candidate start indices not yet selected as centroids. 
One new centroid is drawn according to this distribution, added to $\mathcal{P}$, and the distances $d_s$ and probabilities $p_s$ are updated accordingly. The procedure repeats until $\lvert \mathcal{P} \rvert = P$.

\paragraph{\textbf{4. Iterative Cluster Generation and Centroid Update}}

With $P$ centroids initialized, SIMPC performs $I$ clustering iterations, each involving greedy segmentation and centroid updates. At each time step $t$, we consider all segment lengths $\ell \in [L_{\min}, L_{\max}]$ and extract candidate segments $\mathbf{q}_{t,\ell} = \tilde{\mathbf{X}}_{t : t + \ell} \in \mathbb{R}^{\ell \times D}$. After per-variable Min-Max normalization, each segment $\hat{\mathbf{q}}_{t,\ell}$ is compared against all centroids, and the one with the minimum DTW distance is selected:
\[
(d^\star, C^\star, \ell^\star) = 
\arg\min_{C \in \mathcal{P},\, \ell} 
d_{\text{DTW}}\left(\hat{\mathbf{q}}_{t,\ell},\, C\right).
\]
If the minimum distance $d^\star$ is less than or equal to the DTW threshold $\delta$, the segment is assigned to $C^\star$. The time index is then advanced by a fixed stride length.
After one full pass over $\tilde{\mathbf{X}}$, clusters with at least $\kappa$ members are recompressed using DBA, while smaller clusters are discarded.
The resulting barycenters are then renormalized via per-variable Min-Max scaling and retained as updated centroids. 
If the number of clusters falls below $P$, the K-means++ initialization (step 3) is repeated to maintain $|\mathcal{P}| = P$,
using the updated centroids as fixed seeds. 
During this step, start indices $s$ that have already been assigned to existing clusters are excluded from consideration.
Finally, upon completing $I$ clustering iterations, any pair of centroids $(C_i, C_j)$ such that $d_{\text{DTW}}(C_i, C_j) \leq \delta$ is merged via DBA, yielding the final centroid set $\mathcal{C} = \{C_1, \dots, C_{P'}\}$ with $P' \leq P$.

\paragraph{\textbf{5. Abridged Pseudocode}}

\begin{algorithm}[t]
\caption{SIMPC}
\label{alg:simpc}
\small  
\begin{algorithmic}[1]
\Require Denoised input $\tilde{\mathbf{X}}\in\mathbb{R}^{T\times D}$; desired cluster count $P$; fixed seeds $\{\mathcal{T}_1, \dots, \mathcal{T}_m\}$; segment-length bounds $(L_{\min}, L_{\max})$; DTW threshold $\delta$; minimum cluster size $\kappa$; number of iterations $I$; stride $\Delta$

\Statex {\scriptsize\texttt{// -----2. Domain Adaptation-----}}
\State $\mathcal{P}_0 \leftarrow \{\mathcal{T}_1, \dots, \mathcal{T}_m\}$

\Statex {\scriptsize\texttt{// -----3. Adapted K-means++ Initialization-----}}
\State $\mathcal{P} \gets \mathcal{P}_0$ \Comment{\textcolor{blue}{$\mathcal{P}$: Current centroid set}}
\While{$|\mathcal{P}| < P$}
    \ForAll{unselected start indices $s$}
        \State  $\hat{\mathbf{q}}_{s,L_{\max}} \gets \text{MinMax}(\tilde{\mathbf{X}}_{s : s + L_{\max}})$
        \State 
        $d_s \gets \min_{C \in \mathcal{P}} d_{\text{DTW}}(\hat{\mathbf{q}}_{s,L_{\max}}, C)$ 
    \EndFor
    \State Sample $s^\star$ with probability $p_{s} = d_{s} / \sum_{i \in \mathcal{S}} d_i$
    \State  $\mathcal{P} \gets \mathcal{P} \cup \{\hat{\mathbf{q}}_{s^\star,L_{\max}}\}$
\EndWhile

\Statex {\scriptsize\texttt{// -----4. Iterative Clustering and Centroid Update-----}}
\For{$\textit{iter} = 1$ \textbf{to} $I$}
    \State $t \gets 1$;\quad initialize empty cluster map $\mathcal{C}$
    \While{$t + L_{\min} \le T$}
        \ForAll{$\ell \in [L_{\min}, L_{\max}]$}
            \State $\hat{\mathbf{q}}_{t,\ell} \gets \text{MinMax}(\tilde{\mathbf{X}}_{t : t + \ell})$
            \State Compute $d(C,\ell) \gets d_{\text{DTW}}(\hat{\mathbf{q}}_{t,\ell}, C)$ for all $C \in \mathcal{P}$
        \EndFor
        \State $(d^\star, C^\star, \ell^\star) \gets \arg\min_{C,\ell} d(C,\ell)$
        \If{$d^\star \le \delta$}
             $\mathcal{C}[C^\star] \cup= \tilde{\mathbf{X}}_{t : t + \ell^\star}$ \EndIf
        \State $t \gets t + \Delta$
    \EndWhile
    \ForAll{$C \in \mathcal{C}$}
        \If{$|\mathcal{C}[C]| \ge \kappa$}
            $C \gets 
            \text{MinMax}(\operatorname{DBA}_{\text{DTW}}(\mathcal{C}[C]))$
        \Else\ remove $C$ from $\mathcal{P}$
        \EndIf
    \EndFor
    \State Repeat lines 3–10 until $|\mathcal{P}| = P$
\EndFor
\State Merge centroids with $d_{\text{DTW}}\le\delta$
\State \Return Final centroid set $\mathcal{C}$
\end{algorithmic}
\end{algorithm}

Algorithm~\ref{alg:simpc} is the SIMPC pseudocode. It outputs a pattern set $\mathcal{C}$, which serves as explicit labels for the subsequent shapelet-based classifier, thereby converting JISC-Net’s pattern detection into a fully
supervised learning task.
\section{JISC-Net: Pattern Detection}

\textit{Joint-variable Invariant Shapelet-Classification Network} (JISC-Net) extends the \textit{ShapeNet}~\cite{Li2021ShapeNet}, which is designed to discover shapelets---short, discriminative subsequences that characterize class-specific patterns in multivariate time series. We adopt this shapelet-based approach due to its inherent interpretability, as it enables identification of motifs that directly contribute to classification decisions, a desirable property in financial applications where model transparency is critical.
ShapeNet operates by sliding variable-length windows across all input channels, embedding the resulting candidates using a causal CNN, and selecting the most class-informative shapelets through end-to-end training. Among existing shapelet-based models, ShapeNet is particularly well-suited to our task, as its ability to learn variable-length shapelets allows it to flexibly adapt to temporal variations and effectively handle the amplitude-scaling effects commonly observed in financial time series. \textbf{JISC-Net differs from the original ShapeNet in two key aspects}:  
\textbf{(1)} We regard the entire $D$-dimensional observation $\mathbf{x}_t$ as a \emph{single token}, rather than processing each channel separately. This preserves cross-variable interactions that are critical in multivariate time series.  
\textbf{(2)} We replace the Euclidean distance used for triplet loss function in ShapeNet with a DTW-based distance, endowing the encoder with robustness to temporal distortion. These changes allow JISC-Net to capture richer inter-variable dynamics while remaining resilient to heterogeneous temporal scales.

\paragraph{\textbf{1. Mdc-CNN Encoder with DTW-based Triplet Loss.}}

\begin{figure*}[!htbp]
    \centering
    \includegraphics[width=0.95\linewidth]{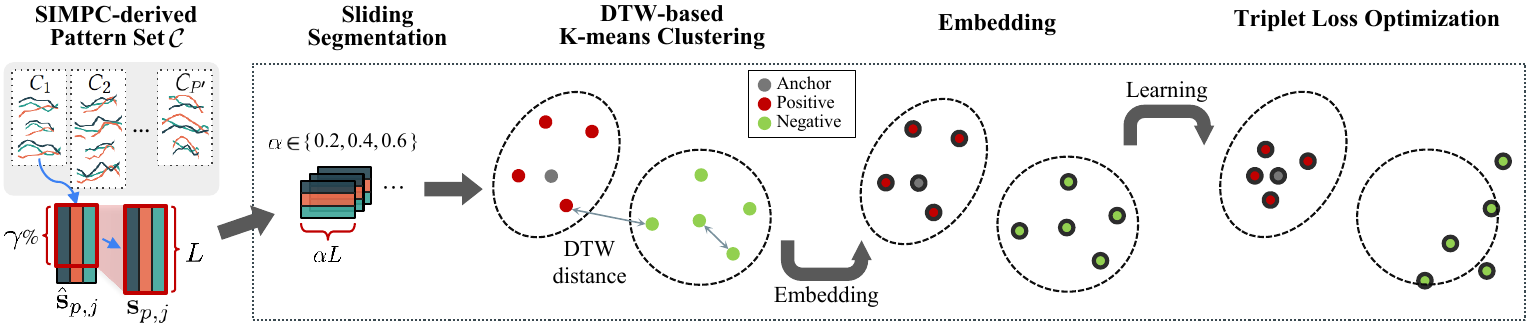} 
    \caption{The encoder learning process of JISC-Net}
    \Description{The encoder learning process of JISC-Net.}
    \label{fig:mdc_cnn}
\end{figure*}

\begin{figure}[!b]
    \centering
    \includegraphics[width=\linewidth]{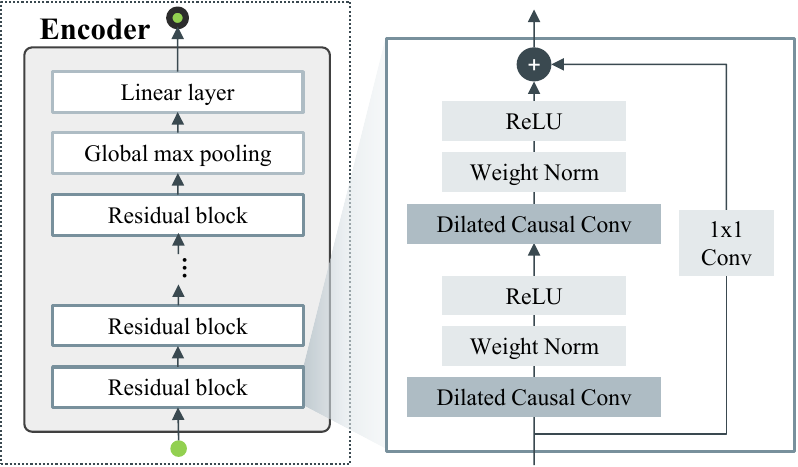} 
    \caption{Architecture of the encoder Mdc-CNN}
    \Description{Architecture of the encoder Mdc-CNN}
    \label{fig:encoder}
\end{figure}

We consider all pattern-labeled subsequences $\mathbf{s}_{p,j} \in C_p$ from clusters $\mathcal{C} = \{ C_1, \ldots, C_{P'} \}$, output by SIMPC. In pattern detection, only the initial $\gamma$ fraction of each subsequence serves as input. 
Figure~\ref{fig:mdc_cnn} shows the encoder training process, and Figure~\ref{fig:encoder} its architecture.
Each sequence is linearly interpolated to a fixed length $L (> \gamma \ell)$ and column-wise Min-Max normalized, yielding $\hat{\mathbf{s}}_{p,j} \in \mathbb{R}^{L \times D}$.
In order to apply the triplet loss function during the encoder training, we need to specify anchor, positive, and negative points. To achieve this, we iterate over the multiscale slide-ratio set $\boldsymbol{\alpha}$, where the elements of $\boldsymbol{\alpha}$ lie between 0 and 1. For each $\alpha \in \boldsymbol{\alpha}$, a transposed subsequence $\hat{\mathbf{s}}_{p,j}^{\top}$ is repeatedly sliced into segments of length $\alpha L$ by sliding a window with a fixed stride. We then cluster the segments for each $\alpha$ using \emph{the dependent DTW-based K-means} (K=2). This multiscale slicing increases training instances and exposes the encoder to local patterns at multiple resolutions. This process applies segmentation and clustering to each $\alpha$ and $\hat{\mathbf{s}}_{p,j}^{\top}$ pair individually, ensuring the encoder receives input shaped as $(-1, D, \alpha L)$.

Let $c$ be one of the two clusters produced by DTW-based K-means. For each of the two clusters, we select
(i) the segment that is located most centrally within the cluster as the \emph{anchor} $\mathbf{A}$,
(ii) the $\lceil |c| / 5 \rceil$ segments closest to the anchor as the \emph{positive set} $S^+$, and
(iii) the $\lceil |\bar{c}| / 5 \rceil$ segments farthest from the anchor, drawn from the opposite cluster $\bar{c}$, as the \emph{negative set} $S^-$.
These selected anchor, positive, and negative points are then embedded through the encoder, and during the optimization process using the triplet loss function, the encoder learns meaningful feature representations.

The Mdc-CNN encoder $f_\theta$ maps each input segment of shape $(D, \alpha L)$ to an embedding $\mathbf{z} \in \mathbb{R}^h$, where
$\mathbf{z}_a, \mathbf{z}_p, \mathbf{z}_n = f_\theta(\mathbf{A}),\; f_\theta(S^+),$\;
$ f_\theta(S^-)$
with masking applied to accommodate the varying sizes of $S^+$ and $S^-$. 
We then compute the distances $d_{ap}$ (anchor-positive), $d_{an}$ (anchor-negative), and intra-class distances $d_{\text{intra}}^{+}$ and $d_{\text{intra}}^{-}$.
\[
\begin{aligned}
d_{ap} &= \frac{1}{|S^+|} \sum_j \|\mathbf{z}_a - \mathbf{z}_{p_j}\|_2, \quad d_{an} = \frac{1}{|S^-|} \sum_j \|\mathbf{z}_a - \mathbf{z}_{n_j}\|_2, \\
d_{\text{intra}}^{+} &= \frac{1}{|S^+|^2} \sum_{i,j} \|\mathbf{z}_{p_i} - \mathbf{z}_{p_j}\|_2, \quad d_{\text{intra}}^{-} = \frac{1}{|S^-|^2} \sum_{i,j} \|\mathbf{z}_{n_i} - \mathbf{z}_{n_j}\|_2.
\end{aligned}
\]
For a given anchor, the loss is computed as:
\[
\ell
= \log \left(
        \frac{d_{ap} + d_{\text{intra}}^{+} + d_{\text{intra}}^{-}}
             {d_{an} + \varepsilon} + m
      \right),
\]
where $\varepsilon = 10^{-6}$ prevents numerical instability and $m = 0.2$ implements a soft margin.
This formulation penalizes large anchor-positive distances and weak intra-cluster cohesion while encouraging anchor-negative separation. 
The total loss is averaged across all segments sampled from $\hat{\mathbf{s}}_{p,j}$.
By replacing Euclidean distance with DTW, the encoder learns invariance to temporal distortions. 
Moreover, all variable channels are encoded jointly, preserving cross-variable dependencies.

\paragraph{\textbf{2. Shapelet Discovery.}}
This phase is dedicated to the discovery of discriminative shapelets using the encoder trained in the preceding phase. The normalized subsequence $\hat{\mathbf{s}}_{p,j}^{\top}$ is segmented at each slide ratio $\alpha$, in accordance with the procedure described earlier.
Each segment is encoded into a fixed-length vector using the trained encoder \(f_{\theta}\).
All embeddings corresponding to segments derived from every $(\hat{\mathbf{s}}_{p,j}^{\top}, \alpha)$ pair are subsequently collected into a latent cloud.
Since temporal distortion invariance has already been learned by $f_\theta$, we apply basic Euclidean K-means clustering (hyperparameter K = $g$) to the latent cloud, yielding $g$ clusters. For each cluster, we select the embedded segment with the smallest Euclidean distance to the cluster center as the shapelet candidate.
The corresponding pre-embedding segment is then traced to recover the raw multivariate curve, denoted as $\mathrm{Shp}_{c} \in \mathbb{R}^{D \times \alpha L}$, 
where $c \in \{1,\dots,g\}$ indexes the shapelet clusters.

To ensure that each of the $g$ clusters represents a class-consistent and meaningful group, we evaluate its \textbf{purity}, which is computed using the embedded segments.
The purity of a cluster is determined by the proportion of segments within the cluster that share the same label. The label refers to the pattern to which the segment belongs, as identified by the output of SIMPC. The purity is calculated as the ratio of the number of segments with the most frequent label in the cluster to the total number of segments in that cluster. A cluster is retained if its purity exceeds a threshold of $\frac{1}{P'}$, where $P'$ is the total number of patterns.
To rank the retained clusters, we define a \textbf{utility score} $U_c$ that quantifies the distinctiveness and representativeness of each shapelet candidate. This score is used solely for ranking and does not affect prediction. For shapelet cluster $c$, the utility score is computed as
\[
U_c = |C_c| \cdot \sum_{c' \ne c} \bigl\| Shp_c - Shp_{c'} \bigr\|_2^2,
\]
where $|C_c|$ is the number of segments in cluster $c$. This formulation favors candidates from larger clusters that are well-separated from others in latent space.

\paragraph{\textbf{3. SVM classifier}}

In this phase, shapelet candidates are used to train a \textit{Support Vector Machine} (SVM) for pattern classification. 
For each shapelet candidate $\mathrm{Shp}_c \in \mathbb{R}^{D \times L_c}$, where $L_c$ denotes the length of the shapelet, we compute the minimum DTW distance between $\mathrm{Shp}_c$ and all segments of the same length in $\hat{\mathbf{s}}_{p,j}^{\top}$. Concretely, we slide a window of length \(L_c\) over \(\hat{\mathbf{s}}_{p,j}^{\top}\), extract each segment \(\hat{\mathbf{s}}_{p,j}^{\top}[t:t+L_c]\), and compute the DTW distance between this segment and the raw shapelet \(\mathrm{Shp}_c\). The final feature vector is 
\[
\begin{aligned}
\Phi\left(\hat{\mathbf{s}}_{p,j}^{\top}\right) = \left[ \min_t \mathrm{DTW}\left( \hat{\mathbf{s}}_{p,j}^{\top}[t:t+L_c],\; \mathrm{Shp}_c \right) \right]_{c=1}^{g} \in \mathbb{R}^{g},
\end{aligned}
\]
where $g$ denotes the number of retained shapelets. The feature vectors from all subsequences $\hat{\mathbf{s}}_{p,j}^{\top}$, along with their corresponding pattern labels provided by SIMPC, are then used to train a linear SVM. The trained classifier outputs a probability distribution $P \in [0, 1]^{P'}$ over the $P'$ pattern types. Each subsequence is classified into the pattern with the highest probability, denoted as $p_{\text{max}}$.

To improve prediction accuracy, we apply filtering during both training and inference. In the \textbf{training stage}, we use the Kolmogorov-Smirnov (K-S) test to identify non-discriminative pattern labels. For each label, we collect \(p_{\text{max}}\) values from correctly and incorrectly classified subsequences, compute cumulative distributions over discretized intervals, and apply the K-S test. Labels with p-values above 0.05 are deemed non-discriminative and treated as non-patterns. Filtering is performed using only training data to prevent information leakage.
In the \textbf{inference stage}, we retain only the top $x\%$ of test subsequences sorted by \(p_{\text{max}}\), filtering out low-confidence predictions as non-patterns.

Note that only the initial $\gamma \ell$ portion of each pattern-classified subsequence $\mathbf{s}_{p,j}^{\top}$ serves as input to the classifier, where $\gamma$ is a tunable hyperparameter. The detection of a pattern match signifies both pattern identification and enables a directional forecast for the remaining $(1 - \gamma)\ell$ portion.

\section{Experiments}

We design our experiments to address four research questions (RQs) on the roles of multivariate input, baseline comparisons, confidence-based filtering, and interpretability within the proposed two-stage framework. Code for our framework is available on GitHub\footnote{\url{https://github.com/kimjuwon-coder/SIMPC-JISCNet}}.

\subsection{Experimental Setup}
\paragraph{\textup{\textbf{Data.}}} We evaluate the framework on two financial domains: cryptocurrency (BTC/USD) and equities (S\&P 500). The equities dataset comprises three high-cap stocks from different S\&P 500 sectors: AAPL (Technology), BRK.B (Financials), and XOM (Energy). Each asset is represented by closing price, trading volume, and relative strength index (RSI). We split the datasets into training (BTC/USD: 2014/01/01--2021/12/31; stocks: 2008/01/01--2018/12/31), validation (BTC/USD: 2022/01/01--2023/01/31; stocks: 2019/01/01--2021/07/30), and testing (BTC/USD: 2024/02/01--2025/05/13; stocks: 2021/08/01--2025/05/13) periods. All models, including baselines, are trained and evaluated using the same input features to ensure fair comparison.

\paragraph{\textup{\textbf{Hyperparameter.}}}
We set the pattern length to approximately 20 days with $L_{\min} = 18$ and $L_{\max} = 22$.
SIMPC extracts $P = 8$ patterns, seeded with $m = 6$ canonical chart prototypes, using a DTW threshold $\delta = 2.3$ and minimum cluster size $\kappa = 40$.
JISC-Net uses the first $\gamma = 0.8$ portion of each subsequence for classification, enabling inference over the remaining about 4 steps.
Each subsequence is linearly interpolated to length $L = 100$.
Multiscale shapelets are generated with sliding ratios $\boldsymbol{\alpha} = \{0.2,\, 0.4,\, 0.6\}$, and the number of shapelets is set to $g = 10$.

\subsection*{RQ1: Effect of Input Variables}
We investigate how different combinations of input variables affect the structure and separability of patterns extracted by the SIMPC algorithm. Our comparison includes: (i) closing price only (C), (ii) closing price and volume (CV), and (iii) closing price, volume, and RSI (CVR). 
Since dependent DTW accumulates Euclidean distances across dimensions, we scale the clustering threshold $\delta$ according to input dimensionality: $\delta = 2.3 / \sqrt{3}$ for 1D inputs, $\delta = 2.3 / \sqrt{2}$ for 2D, and $\delta = 2.3$ for 3D. 
Table~\ref{tab:rq1_dtw} summarizes the number of subsequences extracted as candidate patterns and the pairwise DTW distances between centroids, before applying the final merging step based on $\delta$. To ensure a fair comparison across input dimensionalities, both the average and minimum centroid-level DTW distances (Avg. Dist., Min. Dist.) are normalized by $\sqrt{d}$. The number of extracted subsequences is comparable across all configurations (C: 658, CV: 657, CVR: 642), with an average of approximately 80 subsequences per cluster. However, clustering quality differs significantly depending on input dimensionality. A larger average DTW distance indicates greater diversity among pattern centroids, and a larger minimum distance implies better separation even between the most similar clusters. In the C-only case, several centroid pairs (e.g., (3, 4), (3, 5), (1, 7)) show distances below the threshold $\delta$, suggesting potential structural overlap.
These results confirm that adding volume and RSI as input variables enhances the distinctiveness of extracted pattern centroids. In particular, the CVR configuration yields the most diverse and well-separated centroids, even before any merging step, highlighting the importance of multivariate inputs in capturing structural variation in financial time series.
\begin{table}[t]
\caption{Clustering outcomes across input configurations.}
\label{tab:rq1_dtw}
\centering
\begin{tabular}{ccccc >{\raggedright\arraybackslash}p{2cm}}
\toprule
\textbf{Input} & \textbf{\# Subseq.} & \textbf{Avg. Dist.} & \textbf{Min. Dist.} & \textbf{Below $\delta$} \\
\midrule
C   & 658 & 1.6428 & 0.2174 & (3,4), (3,5), (1,7) \\
CV  & 657 &1.698 &0.759 & None \\
CVR & 642 &1.743 & 0.898 & None \\
\bottomrule
\end{tabular}

\end{table}

\subsection*{RQ2: Baseline Comparison}


\begin{table*}[!ht]
\renewcommand{\arraystretch}{0.85}
\centering
\caption{Evaluation metrics for classification and trading performance. \textbf{Bold} and \underline{underline} indicate the best and second-best results among baselines and Ours-T@100.
The \protect\highlightbox{red box} highlights the overall best performance in WLR, AR, and TRwf.}
\label{tab:metrics_btc_aapl_notrwof}

\small
\begin{tabularx}{\textwidth}{
  l
  |>{\centering\arraybackslash}X>{\centering\arraybackslash}X>{\centering\arraybackslash}X>{\centering\arraybackslash}X|>{\centering\arraybackslash}X
  |>{\centering\arraybackslash}X>{\centering\arraybackslash}X>{\centering\arraybackslash}X>{\centering\arraybackslash}X|>{\centering\arraybackslash}X
}
\toprule
\textbf{Methods}
& \multicolumn{5}{c|}{\textbf{BTC/USD}} 
& \multicolumn{5}{c}{\textbf{AAPL}} \\
\cmidrule(lr){2-6} \cmidrule(lr){7-11}
& F1$^{\uparrow}$ & WLR$^{\uparrow}$ & AR$^{\uparrow}$ & TRwf$^{\uparrow}$ & \#Trades
& F1$^{\uparrow}$ & WLR$^{\uparrow}$ & AR$^{\uparrow}$ & TRwf$^{\uparrow}$ & \#Trades \\
\midrule
Random        & 0.478 & 0.937 & -0.256 & -0.842 & 185  & 0.519 & 1.073 & -0.016 & -0.249 &115 \\
\midrule
LightGBM   & 0.392 & 1.229 & 0.823 & 1.152 &   & 0.368 & 1.130 & 0.345  & 0.167 & \\
DoubleEnsemble   & 0.267 & 0.779 & -0.819 & -1.885 &   & 0.309 & 0.917 & -0.090  & -0.334 & \\
ADARNN   & 0.398 & \underline{1.256} & \underline{0.966} & \underline{1.417} &   & 0.358 & 1.091 & 0.078  & -0.141 & \\
ALSTM   & 0.493 & 1.056 & 0.590 & 0.721 & 185 & 0.358 & 1.091 & 0.320  & 0.138 & 115\\
TCTS   & 0.386 & 1.202 & 0.920 & 1.332 &   & 0.348& 1.054 & 0.065  & -0.156 & \\
Tabnet   & 0.398 & \underline{1.256} & 0.897 & 1.289 &  & 0.338 & 1.018 & -0.129  & -0.379 & \\
Transformer   & 0.487 & 0.983 & -0.113 & -0.359 &     & 0.487 & 0.983 & -0.113  & -0.359 & \\
\midrule
DLinear       & 0.511 & 1.102 & 0.399 & 0.369 &     & \underline{0.513} & 1.396 & \underline{0.499}  & \underline{0.343} &  \\
Autoformer    & 0.490 & 1.091 & -0.017 & -0.249 &     & 0.490 & 1.091 & -0.017 & -0.249 &  \\
FEDformer     & 0.497 & 1.229 & 0.217 & 0.031 &    & 0.401 & 0.983 & 0.380  & 0.207 &  \\
Informer      & \underline{0.532} & 1.056 & 0.523 & 0.598 & 185   & 0.458 & 1.255 & 0.211  & 0.013 & 115 \\
iTransformer  & 0.456 & 1.079 & -0.215 & -0.768 &     & 0.460 & 1.054 & 0.292  & 0.106 &  \\
PatchTST      & 0.498 & \underline{1.256} & 0.522 & 0.597 &     & 0.453 & 1.396 & 0.472  & 0.312 &  \\
TimesNet      & 0.466 & 1.033 & 0.322 & 0.226 &     & 0.464 & \underline{1.500} & 0.420  & 0.253 &  \\
\midrule
\rowcolor{gray!30}
Ours-T@100    & \textbf{0.594} & \textbf{1.534} & \textbf{1.066} & \highlightbox{\textbf{1.602}} & 185  & \textbf{0.541} & \textbf{1.614} & \textbf{0.602} & \highlightbox{\textbf{0.462}} & 115 \\
Ours-T@80     & 0.591 & \highlightbox{1.561} & 1.261 & 1.549 & 146   & 0.506 & 1.583 & 0.557 & 0.332 & 93 \\
Ours-T@60     & 0.567 & 1.458 & 1.265 & 1.257 & 118   & 0.543 & 2.000 & 0.688 & 0.381 & 78 \\
Ours-T@40     & 0.511 & 1.452 & \highlightbox{1.576} & 1.046 & 76   & 0.565 & 2.375 & 0.747 & 0.295 & 54 \\
Ours-T@20     & 0.407 & 1.375 & 1.196 & 0.378 & 38   & 0.550 & \highlightbox{2.429} & \highlightbox{1.166} & 0.232 & 24 \\
\bottomrule
\end{tabularx}

\centering
\small
\begin{tabularx}{\textwidth}{
  l
  |>{\centering\arraybackslash}X>{\centering\arraybackslash}X>{\centering\arraybackslash}X>{\centering\arraybackslash}X|>{\centering\arraybackslash}X
  |>{\centering\arraybackslash}X>{\centering\arraybackslash}X>{\centering\arraybackslash}X>{\centering\arraybackslash}X|>{\centering\arraybackslash}X
}
\toprule
\textbf{Methods}
& \multicolumn{5}{c|}{\textbf{BRK.B}} 
& \multicolumn{5}{c}{\textbf{XOM}} \\
\cmidrule(lr){2-6} \cmidrule(lr){7-11}
& F1$^{\uparrow}$ & WLR$^{\uparrow}$ & AR$^{\uparrow}$ & TRwf$^{\uparrow}$ & \#Trades
& F1$^{\uparrow}$ & WLR$^{\uparrow}$ & AR$^{\uparrow}$ & TRwf$^{\uparrow}$ & \#Trades \\
\midrule
Random        & 0.544 & 1.146 & -0.043 & -0.539 & 222   & 0.523 & 1.075 & 0.094 & -0.235 & 221 \\
\midrule
LightGBM   & 0.434 & 1.220 & 0.269 & 0.153 &  & 0.499 & 1.210 & 0.429  & 0.506 &  \\
DoubleEnsemble   & 0.261 & 0.748 & -0.282 & -1.070 &  & 0.282 & 0.811 & -0.288 & -1.079 &  \\
ADARNN   & 0.411 & 1.312 & 0.308 & 0.240 &   & 0.403 & 1.278 & 0.324 & 0.275 & \\
ALSTM   & 0.406 & 1.289 & 0.295 & 0.210 & 222 & 0.393 & 1.232 & 0.300 & 0.221 & 221 \\
TCTS   & 0.422 & 1.362 & 0.297 & 0.215 &  & 0.425 & 1.376 & 0.405  & 0.454 & \\
Tabnet   & 0.443 & 1.337 & \underline{0.344} & \underline{0.320} & & 0.502 & 1.255 & 0.323 & 0.271 & \\
Transformer   & 0.492 & 1.413 & 0.256 & 0.125 & & 0.495 & \textbf{1.429} & \textbf{0.756} & \textbf{1.228} &\\
\midrule
DLinear       & \underline{0.597} & 1.056 & -0.002 & -0.448 &   & 0.537 & 1.376 & 0.590 & 0.861 & \\
Autoformer    & 0.581 & 1.220 & 0.088 & -0.248 &    & \underline{0.556} & 0.973 & 0.153 & -0.104 & \\
FEDformer     & 0.591 & 1.440 & 0.217 & 0.037 &     & 0.530 & 1.085 & 0.153 & -0.103 & \\
Informer      & 0.568 & 1.289 & 0.218 & 0.041 & 222   & 0.545 & 1.376 & 0.535 & 0.740 & 221 \\
iTransformer  & 0.566 & \textbf{1.523} & 0.253 & 0.117 &     & 0.521 & 1.232 & 0.467 & 0.590 & \\
PatchTST      & \textbf{0.601} & 1.155 & 0.064 & -0.302 &    & 0.540 & 1.146 & -0.024 & -0.494 & \\
TimesNet      & 0.550 & 1.312 & 0.296 & 0.213 &  & 0.534 & 1.255 & 0.448 & 0.547 & \\
\midrule
\rowcolor{gray!30}
Ours-T@100    & 0.573 & \highlightbox{\underline{1.467}} & \textbf{0.608} & \highlightbox{\textbf{0.907}} & 222  & \textbf{0.573} & \underline{1.402} & \underline{0.593} & \underline{0.868} & 221 \\
Ours-T@80     & 0.561 & 1.392 & 0.543 & 0.608 & 177  & 0.606 & 1.603 & 0.889 & \highlightbox{1.219} & 177 \\
Ours-T@60     & 0.550 & 1.315 & 0.647 & 0.559 & 125   & 0.625 & 1.698 & 1.030 & 1.187 & 143 \\
Ours-T@40     & 0.544 & 1.289 & 0.556 & 0.310 & 87   & 0.624 & 1.667 & 1.005 & 0.708 & 88 \\
Ours-T@20     & 0.545 & 1.316 & \highlightbox{0.681} & 0.212 & 44   & 0.723 & \highlightbox{2.625} & \highlightbox{1.567} & 0.396 & 29 \\
\bottomrule
\end{tabularx}

\end{table*}

\paragraph{\textup{\textbf{Evaluation Metrics.}}} We report four metrics~\cite{surveyfinancialaiarchitectures}: F1-score (F1), Win-Loss Ratio (WLR), Average Return (AR), and Total Return with fees (TRwf). F1 captures classification performance under class imbalance. WLR measures the ratio of correctly predicted directional trades to incorrect ones, while AR quantifies the expected profitability per trade. TRwf indicates cumulative returns after transaction costs. All metrics except TRwf ignore fees.

\paragraph{\textup{\textbf{Trading Protocol.}}}
All experiments follow a representative trading protocol~\cite{ijcai2024p678, zhang2020doubleensemble} designed to reflect realistic market conditions. Every 4 days during the test period, a position---long or short---is opened based on the predicted 4-day movement and closed at the end of that window. Each trade is executed with a fixed amount, and a 0.1\% transaction fee is applied at both entry and exit. To ensure fair comparison, all models are restricted to the same number of trades as our method, preventing frequency-related distortions in return-based metrics. For all baselines except Random, trades are executed only when the prediction confidence is among the top $k$ at fixed 4-day intervals, where $k$ equals the number of trades in our method.
For the Random strategy, we randomly select $k$ trading points from the same 4-day intervals to match the number of trades. 
This ensures that all models operate under equivalent market exposure and decision frequency, while accounting for prediction confidence.

\paragraph{\textup{\textbf{Baselines.}}}
We compare our framework against three baseline categories\cite{ijcai2024p678}: 
(1) \textbf{Random}, 
(2) \textbf{Quantitative Investment Methods} frequently employed in Microsoft's open-source quantitative research platform Qlib~\cite{yang2020qlib}: LightGBM\cite{ke2017lightgbm}, DoubleEnsemble\cite{zhang2020doubleensemble}, ADARNN\cite{du2021adarnn}, ALSTM\cite{qin2017alstm}, TCTS\cite{wu2021tcts}, TabNet\cite{arik2019tabnet}, Transformer\cite{vaswani2017attention}, and (3) a recent \textbf{Time-series Methods}: DLinear\cite{zeng2022dlinear}, Autoformer\cite{wu2021autoformer}, FEDformer\cite{zhou2022fedformer}, Informer\cite{zhou2021informer}, iTransformer\cite{liu2024itransformer}, PatchTST\cite{Nie2023PatchTST}, and TimesNet\cite{wu2023timesnet}.

\paragraph{\textup{\textbf{Performance Analysis.}}}
Table~\ref{tab:metrics_btc_aapl_notrwof} summarizes the binary classification and trading outcomes on four assets. We denote \texttt{Ours-T@x} as our method with top-$x\%$ confidence filtering during inference. 
Without confidence filtering (Ours-T@100), our method achieves the \textbf{best performance} in 8 out of 12 metric–asset combinations and ranks \underline{second-best} in 3 of the remaining 4, considering the primary metrics (F1, WLR, AR) while excluding TRwf due to its redundancy with AR.
Regarding profitability (AR), Ours-T@100 records the highest average return on three assets and the second highest on the fourth. The fee-adjusted total return (TRwf) also remains positive and maximal, demonstrating practical economic value even under conservative cost assumptions. 
Our model and the baselines execute the same number of trades, so the performance improvement does not arise from reducing trade frequency or selective execution, but from genuinely superior signal quality. 
These advantages come from the architectural properties of our model. The proposed framework postpones prediction until a recurring pattern with temporal distortion is reliably detected, allowing the model to bypass high-noise segments that typically mislead baseline models. By focusing on a limited, explainable subset of time steps with high predictive confidence, rather than continuously issuing low-confidence signals, the framework demonstrates statistically and economically significant improvements in noisy financial environments.

\subsection*{RQ3: Two-Stage Filtering in JISC-Net}

We analyze the two-stage filtering process in JISC-Net that operates on the output of the SVM-based pattern classifier. During training, it filters pattern labels with low statistical separability via the Kolmogorov-Smirnov (K-S) test, and during inference, it imposes a dynamic threshold to retain only high-confidence predictions.

\paragraph{\textup{\textbf{Stage 1: Per-label filtering based on the K-S test.}}}
We first apply the K-S test to each pattern label to evaluate whether prediction confidence can effectively distinguish correct from incorrect classifications. A label is retained only if the p-value of the K-S test is below 0.05. Table~\ref{tab:ks_pvalues} presents the p-values for XOM labels, revealing that label 7 is the only one that failed the test and was reclassified as a non-pattern.
Applying the K-S filter removes 13.90\% of test subsequences in BTC/USD, 41.47\% in AAPL, 8.74\% in BRK.B, and 5.60\% in XOM, corresponding to subsequences associated with pattern labels that exhibit low separability.

\begin{table}[H]
\centering
\caption{K–S test p-values per label (XOM dataset).}
\label{tab:ks_pvalues}
\resizebox{0.48\textwidth}{!}{%
\begin{tabular}{cccccccccc}
\toprule
\textbf{Label} & 0 & 1 & 2 & 3 & 4 & 5 & 6 & 7 \\
\midrule
\textbf{p-value} & $1.98\mathrm{e}{-2}$ & $1.00\mathrm{e}{-4}$ & $5.00\mathrm{e}{-4}$ & $2.16\mathrm{e}{-10}$ & $1.32\mathrm{e}{-6}$ & $1.76\mathrm{e}{-2}$ & $1.93\mathrm{e}{-3}$ & $7.05\mathrm{e}{-1}$ \\
\bottomrule
\end{tabular}
}

\end{table}

\paragraph{\textbf{\textup{Stage 2: Global thresholding based on confidence.}}}
In the second stage, we sort all predictions by their maximum class probability and retain only the top $x\%$ for final inference. Table~\ref{tab:metrics_btc_aapl_notrwof} compares Ours-T@100, T@80, T@60, and T@20 using win-loss ratio (WLR), average return (AR), and fee-adjusted total return (TR$\text{wf}$). We observe that moderately aggressive thresholds provide a favorable balance between return and risk. For instance, on XOM, T@80 achieves a strong AR (0.889) and the highest TR$\text{wf}$ (1.219) across all variants, while maintaining a reasonable number of trades (177). This indicates that filtering out low-confidence predictions can improve profitability without significantly reducing coverage. In contrast, more aggressive filtering (e.g., T@20 on XOM) yields the highest AR (1.567) but significantly reduces profitability (0.396) due to sparse trades (29).

\begin{figure}[t]
    \centering
    \includegraphics[width=\linewidth]{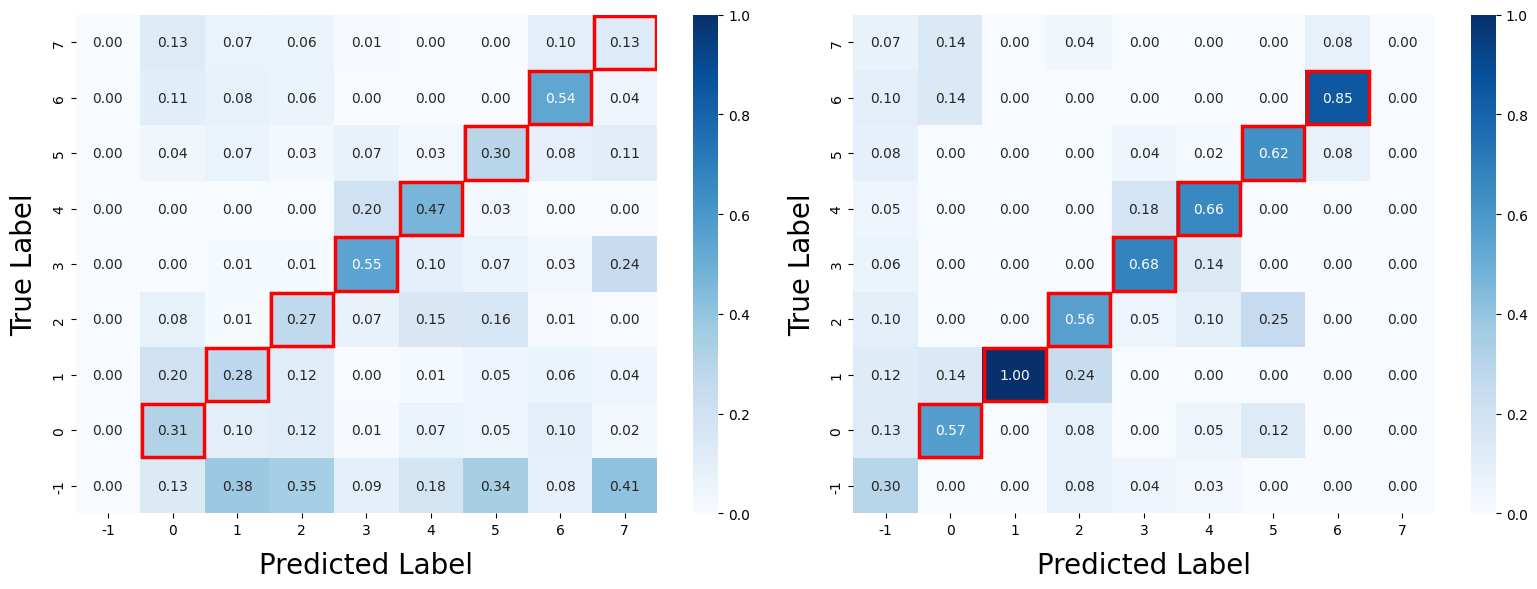}
    \caption{Column-normalized confusion matrices for XOM before (Left) and after (Right, T@30) two-stage filtering. Red boxes mark correct predictions; label -1 indicates non-patterns assignments.}
    \label{fig:xom_confusion}
    \Description{Two confusion matrices for XOM. The left matrix shows model output before filtering, and the right shows output after applying temporal filtering at T=30. The red cells indicate model-classified patterns. After filtering, label 7 disappears, meaning it's treated as non-pattern. Label -1 is used for non-pattern assignments.}
\end{figure}

\paragraph{\textbf{\textup{Qualitative impact on confusion structure.}}}
Figure~\ref{fig:xom_confusion} illustrates the effect of the two-stage filtering pipeline on classification consistency. The confusion matrix is normalized column-wise to reflect the distribution of true class assignments for each predicted label.
Without any filtering, the model frequently misclassifies instances across classes, including assigning noise (label -1) to pattern classes and vice versa.
After applying the K-S filtering, label 7---whose p-value exceeded 0.05 in the K-S test (Table~\ref{tab:ks_pvalues})---is treated entirely as noise, reducing some misclassifications.
Following the top $x\%$ thresholding (T@30), the confusion matrix becomes more distinctly diagonal, indicating that the majority of noisy sequences are confidently rejected and pattern ambiguity is further minimized.
Overall, the two-stage filtering substantially improves both prediction precision and reliability in pattern classification.

\subsection*{RQ4: Pattern-Based Prediction Analysis}

\begin{figure}[!t]
    \centering
    \includegraphics[width=\linewidth]{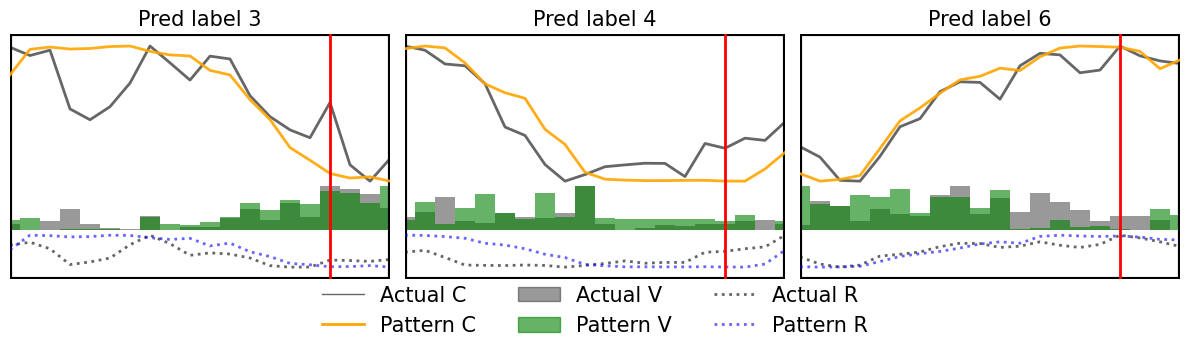}
    \caption{Successful prediction cases. The vertical red line marks the 16th ($\gamma = 0.8$) day, when trading is triggered.}
    \label{fig:correct}
    \Description{A line chart showing several time series segments that align closely with predicted patterns. Each example highlights a clear match between the input sequence and a known pattern. A vertical red line marks the 16th day, the point at which a trade is executed.}
\end{figure}

\begin{figure}[ht]
    \centering
    \includegraphics[width=\linewidth]{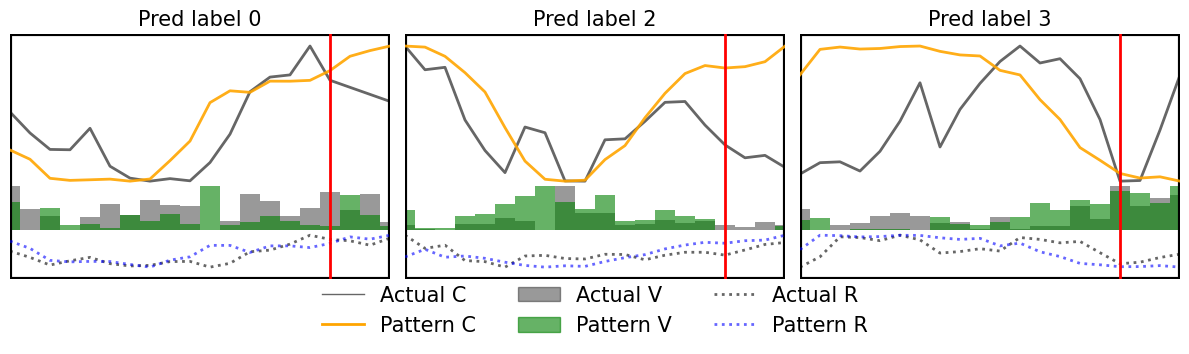}
    \caption{Failure prediction cases.}
    \label{fig:incorrect}
    \Description{A line chart showing multiple time series examples where the actual price movement deviates significantly from the predicted pattern. The misalignment leads to failed trade outcomes, illustrating cases where the model's pattern-based prediction did not match reality.}
\end{figure}

To interpret the model’s decision-making process, we conduct a case study on the XOM asset by analyzing several instances of both successful and failed trades executed by JISC-Net. For each case, we examine: (i) the input sequence used for classification (the initial 16 time steps), (ii) the corresponding pattern predicted by JISC-Net, and (iii) the subsequent 4-step actual price trajectory following the trade.
Figure~\ref{fig:correct} presents \textbf{three success cases} in which the model correctly predicted the directional movement, resulting in profitable trades. In all examples, we observe strong alignment between the input sequence and the predicted pattern across all input variables. For instance, in the first case (predicted as pattern 3), the input sequence exhibits a steep decline in price, a surge in trading volume, and a gradually falling RSI. The predicted pattern mirrors these characteristics and projects a further mild decrease in price, sustained high volume, and persistently low RSI in the following 4 steps. The actual market trajectory closely follows this projection, indicating that the model successfully anticipated the short-term price behavior.
In contrast, Figure~\ref{fig:incorrect} shows \textbf{three failure cases} where the model’s prediction led to losses. In the first two cases (predicted as pattern 0 and pattern 2), although the initial portion of the input sequence aligns moderately well with the predicted pattern, the subsequent market movement deviates significantly from the anticipated trajectory, especially within the 4-day trading window. This suggests that the model may have overfit to the early segment of the input and failed to account for the variability that arises beyond the classification window. One possible remedy is to incorporate probabilistic forecasts of future trajectories rather than relying solely on deterministic directional predictions. Probabilistic modeling can better capture the uncertainty in future price evolution, enabling the model to express multiple plausible outcomes and reduce overconfidence in a single path, thereby mitigating overfitting to specific historical shapes. Exploring this direction could form the basis of future research.
The third failure case demonstrates a clear mismatch even at the initial stage. Although the model assigns this instance to pattern 3, the input sequence diverges substantially from the structure of that pattern. This misclassification likely stems from the limitations of Dynamic Time Warping (DTW), which can occasionally yield artificially low distances when aligning sequences that are temporally shifted but structurally dissimilar. To address this, DTW computations could be regularized by incorporating shape-aware penalties such as constraints on local slope or segment-wise curvature, and by introducing local feature consistency checks that penalize misalignment in critical features (e.g., reversal points, peaks, and troughs). These enhancements would encourage more semantically meaningful alignments and reduce spurious pattern matches.

Overall, these visualizations underscore the interpretability of the proposed framework. Unlike black-box models, JISC-Net combined with SIMPC offers explicit insight into the pattern structure that underlies each trading decision. This not only enhances user understanding but also supports human-in-the-loop decision-making, by allowing users to verify the plausibility of a pattern match before executing a trade.

\balance
\section{Conclusion}

We propose a two-stage framework for pattern-based directional prediction in financial time series. The method integrates SIMPC, a DTW-based unsupervised clustering approach that discovers recurring multivariate patterns with dynamic window sizes under amplitude scaling and temporal distortion, with JISC-Net, a shapelet-based classifier that makes interpretable predictions using partial input sequences. Experiments across four asset classes demonstrate that our method consistently outperforms both quantitative investment models and recent deep learning approaches in directional accuracy and trading performance, while also improving interpretability to support human-in-the-loop decision-making. As future work, we aim to explore probabilistic forecasting and shape-aware alignment to further enhance robustness and generalization.

\begin{acks}
This work was supported by the National Research Foundation of Korea (NRF) grant funded by the Korea government (MSIT) (No. RS-2024–00456247, No. RS-2023–00218913) and by Institute of Information \& communications Technology Planning \& Evaluation (IITP) grant funded by the Korea government (MSIT) (No. RS-2025-25443718, Next-HCAI Project) and (No.RS-2019-II191906, Artificial Intelligence Graduate School Program (POSTECH).
\end{acks}

\appendix
\section{Traditional Stock Chart Patterns}

\label{appendix:chart-patterns}
\begin{figure}[htbp]
    \centering \includegraphics[width=\linewidth]{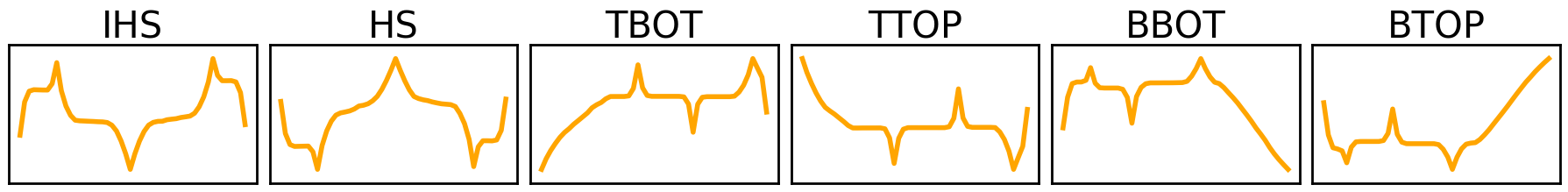}
    \caption{Six initial \textit{price} patterns used as centroids in SIMPC.}
    \label{fig:traditional}
    \Description{An image showing six line plot patterns used as initial centroids in the SIMPC algorithm. These include common financial motifs such as head and shoulders, inverse head and shoulders, and other symmetrical or trend-based shapes.}
\end{figure}

To extract traditional patterns, we utilize historical daily price data from 418 companies that were constituents of the S\&P 500 index at any time between January 2008 and July 2021. We base our pattern definitions on the ten canonical forms introduced by Lo et al.~\cite{lo2000foundations}, which include: Head-and-Shoulders (HS), Inverse Head-and-Shoulders (IHS), Broadening Tops (BTOP), Broadening Bottoms (BBOT), Triangle Tops (TTOP), Triangle Bottoms (TBOT), Rectangle Tops (RTOP), Rectangle Bottoms (RBOT), Double Tops (DTOP), and Double Bottoms (DBOT). These canonical forms are \textbf{rule-based} and defined solely using price movements.
To smooth the price series and reduce short-term volatility, we apply kernel regression using a Gaussian kernel with bandwidth $h = 0.3$.
For peak and trough detection, we first identify local maxima and minima from the smoothed series using a broader kernel ($h = 0.8$) to account for noise. Around each detected extremum, we refine its position by searching for the maximum or minimum value within a $\pm2$ day range of the smoothed signal. Each candidate pattern is then evaluated based on the structural constraints defined in~\cite{lo2000foundations}. For example, the Head-and-Shoulders (HS) pattern consists of five consecutive extremal points where the middle peak is the highest, the two shoulders are lower and roughly symmetrical, and the outer troughs and inner troughs differ by no more than 3\% of the average of the outer peaks.
To capture a wide range of temporal resolutions suitable for the SIMPC algorithm, we consider all window sizes between 15 and 35 trading days. Based on this setting, we identify the most frequently occurring patterns with more than 1,000 instances in the S\&P 500 dataset as follows: IHS (2,567), HS (2,525), TBOT (2,237), BTOP (1,718), TTOP (1,485), and BBOT (1,179).
As shown in Fig.~\ref{fig:traditional}, these six patterns are used as the initial centroids in our SIMPC.

\section*{GenAI Usage Disclosure}
During the preparation of this paper, we used generative AI for grammar and style checking of the English abstract.
No generative AI tools were used in designing experiments or writing the main content of the paper.

\bibliographystyle{ACM-Reference-Format}
\balance
\bibliography{sample-base}


\begin{thebibliography}{36}


\ifx \showCODEN    \undefined \def \showCODEN     #1{\unskip}     \fi
\ifx \showISBNx    \undefined \def \showISBNx     #1{\unskip}     \fi
\ifx \showISBNxiii \undefined \def \showISBNxiii  #1{\unskip}     \fi
\ifx \showISSN     \undefined \def \showISSN      #1{\unskip}     \fi
\ifx \showLCCN     \undefined \def \showLCCN      #1{\unskip}     \fi
\ifx \shownote     \undefined \def \shownote      #1{#1}          \fi
\ifx \showarticletitle \undefined \def \showarticletitle #1{#1}   \fi
\ifx \showURL      \undefined \def \showURL       {\relax}        \fi
\providecommand\bibfield[2]{#2}
\providecommand\bibinfo[2]{#2}
\providecommand\natexlab[1]{#1}
\providecommand\showeprint[2][]{arXiv:#2}

\bibitem[Arik and Pfister(2019)]%
        {arik2019tabnet}
\bibfield{author}{\bibinfo{person}{Sercan~O. Arik} {and} \bibinfo{person}{Tomas
  Pfister}.} \bibinfo{year}{2019}\natexlab{}.
\newblock \bibinfo{title}{TabNet: Attentive Interpretable Tabular Learning}.
\newblock
\showeprint[arxiv]{1908.07442}~[cs.LG]
\newblock
\shownote{Preprint}.


\bibitem[Berndt and Clifford(1994)]%
        {berndt1994using}
\bibfield{author}{\bibinfo{person}{Donald~J Berndt} {and}
  \bibinfo{person}{James Clifford}.} \bibinfo{year}{1994}\natexlab{}.
\newblock \showarticletitle{Using dynamic time warping to find patterns in time
  series}. In \bibinfo{booktitle}{\emph{Proceedings of the 3rd international
  conference on knowledge discovery and data mining}}.
  \bibinfo{pages}{359--370}.
\newblock


\bibitem[Bostrom and Bagnall(2017)]%
        {Bostrom2017Binary}
\bibfield{author}{\bibinfo{person}{Aaron Bostrom} {and}
  \bibinfo{person}{Anthony Bagnall}.} \bibinfo{year}{2017}\natexlab{}.
\newblock \showarticletitle{Binary shapelet transform for multiclass time
  series classification}.
\newblock \bibinfo{journal}{\emph{Transactions on Large-Scale Data-and
  Knowledge-Centered Systems XXXII: Special Issue on Big Data Analytics and
  Knowledge Discovery}} (\bibinfo{year}{2017}), \bibinfo{pages}{24--46}.
\newblock


\bibitem[Du et~al\mbox{.}(2021)]%
        {du2021adarnn}
\bibfield{author}{\bibinfo{person}{Yuntao Du}, \bibinfo{person}{Jindong Wang},
  \bibinfo{person}{Wenjie Feng}, \bibinfo{person}{Sinno Pan},
  \bibinfo{person}{Tao Qin}, \bibinfo{person}{Renjun Xu}, {and}
  \bibinfo{person}{Chongjun Wang}.} \bibinfo{year}{2021}\natexlab{}.
\newblock \showarticletitle{AdaRNN: Adaptive Learning and Forecasting of Time
  Series}.
\newblock \bibinfo{journal}{\emph{CIKM}} (\bibinfo{year}{2021}).
\newblock


\bibitem[Dukkipati et~al\mbox{.}(2025)]%
        {choudhary2024predictive}
\bibfield{author}{\bibinfo{person}{Ambedkar Dukkipati}, \bibinfo{person}{Kawin
  Mayilvaghanan}, \bibinfo{person}{Naveen~Kumar Pallekonda},
  \bibinfo{person}{Sai~Prakash Hadnoor}, {and} \bibinfo{person}{Ranga~Shaarad
  Ayyagari}.} \bibinfo{year}{2025}\natexlab{}.
\newblock \showarticletitle{Predictive AI with External Knowledge Infusion for
  Stocks}.
\newblock \bibinfo{journal}{\emph{arXiv preprint arXiv:2504.20058}}
  (\bibinfo{year}{2025}).
\newblock


\bibitem[Esling and Agon(2012)]%
        {esling2012time}
\bibfield{author}{\bibinfo{person}{Philippe Esling} {and}
  \bibinfo{person}{Carlos Agon}.} \bibinfo{year}{2012}\natexlab{}.
\newblock \showarticletitle{Time-series data mining}.
\newblock \bibinfo{journal}{\emph{ACM Computing Surveys (CSUR)}}
  \bibinfo{volume}{45}, \bibinfo{number}{1} (\bibinfo{year}{2012}),
  \bibinfo{pages}{1--34}.
\newblock


\bibitem[Fan and Shen(2024)]%
        {fan2024stockmixer}
\bibfield{author}{\bibinfo{person}{Jinyong Fan} {and} \bibinfo{person}{Yanyan
  Shen}.} \bibinfo{year}{2024}\natexlab{}.
\newblock \showarticletitle{StockMixer: a simple yet strong MLP-based
  architecture for stock price forecasting}. In
  \bibinfo{booktitle}{\emph{Proceedings of the AAAI Conference on Artificial
  Intelligence}}, Vol.~\bibinfo{volume}{38}. \bibinfo{pages}{8389--8397}.
\newblock


\bibitem[Grabocka et~al\mbox{.}(2014)]%
        {Grabocka2014}
\bibfield{author}{\bibinfo{person}{Josif Grabocka}, \bibinfo{person}{Nicolas
  Schilling}, \bibinfo{person}{Martin Wistuba}, {and} \bibinfo{person}{Lars
  Schmidt-Thieme}.} \bibinfo{year}{2014}\natexlab{}.
\newblock \showarticletitle{Learning time-series shapelets}. In
  \bibinfo{booktitle}{\emph{Proceedings of the 20th ACM SIGKDD international
  conference on Knowledge discovery and data mining}}.
  \bibinfo{pages}{392--401}.
\newblock


\bibitem[Huang et~al\mbox{.}(2024)]%
        {Huang2024FTSDiffusion}
\bibfield{author}{\bibinfo{person}{Hongbin Huang}, \bibinfo{person}{Minghua
  Chen}, {and} \bibinfo{person}{Xiao Qiao}.} \bibinfo{year}{2024}\natexlab{}.
\newblock \showarticletitle{Generative learning for financial time series with
  irregular and scale-invariant patterns}. In \bibinfo{booktitle}{\emph{The
  Twelfth International Conference on Learning Representations}}.
\newblock


\bibitem[Ke et~al\mbox{.}(2017)]%
        {ke2017lightgbm}
\bibfield{author}{\bibinfo{person}{Guolin Ke}, \bibinfo{person}{Qi Meng},
  \bibinfo{person}{Thomas Finley}, \bibinfo{person}{Taifeng Wang},
  \bibinfo{person}{Wei Chen}, \bibinfo{person}{Weidong Ma},
  \bibinfo{person}{Qiwei Ye}, {and} \bibinfo{person}{Tie-Yan Liu}.}
  \bibinfo{year}{2017}\natexlab{}.
\newblock \showarticletitle{LightGBM: A Highly Efficient Gradient Boosting
  Decision Tree}. In \bibinfo{booktitle}{\emph{Advances in Neural Information
  Processing Systems}}.
\newblock


\bibitem[Lee et~al\mbox{.}(2022)]%
        {lee2022tilde}
\bibfield{author}{\bibinfo{person}{Hyunwook Lee}, \bibinfo{person}{Chunggi
  Lee}, \bibinfo{person}{Hongkyu Lim}, {and} \bibinfo{person}{Sungahn Ko}.}
  \bibinfo{year}{2022}\natexlab{}.
\newblock \showarticletitle{TILDE-Q: a transformation invariant loss function
  for time-series forecasting}.
\newblock \bibinfo{journal}{\emph{arXiv preprint arXiv:2210.15050}}
  (\bibinfo{year}{2022}).
\newblock


\bibitem[Li et~al\mbox{.}(2021)]%
        {Li2021ShapeNet}
\bibfield{author}{\bibinfo{person}{Guozhong Li}, \bibinfo{person}{Byron Choi},
  \bibinfo{person}{Jianliang Xu}, \bibinfo{person}{Sourav~S Bhowmick},
  \bibinfo{person}{Kwok-Pan Chun}, {and} \bibinfo{person}{Grace Lai-Hung
  Wong}.} \bibinfo{year}{2021}\natexlab{}.
\newblock \showarticletitle{Shapenet: A shapelet-neural network approach for
  multivariate time series classification}. In
  \bibinfo{booktitle}{\emph{Proceedings of the AAAI conference on artificial
  intelligence}}, Vol.~\bibinfo{volume}{35}. \bibinfo{pages}{8375--8383}.
\newblock


\bibitem[Li et~al\mbox{.}(2023)]%
        {li2023pen}
\bibfield{author}{\bibinfo{person}{Shuqi Li}, \bibinfo{person}{Weiheng Liao},
  \bibinfo{person}{Yuhan Chen}, {and} \bibinfo{person}{Rui Yan}.}
  \bibinfo{year}{2023}\natexlab{}.
\newblock \showarticletitle{PEN: prediction-explanation network to forecast
  stock price movement with better explainability}. In
  \bibinfo{booktitle}{\emph{Proceedings of the AAAI Conference on Artificial
  Intelligence}}, Vol.~\bibinfo{volume}{37}. \bibinfo{pages}{5187--5194}.
\newblock


\bibitem[Li et~al\mbox{.}(2024)]%
        {li2024master}
\bibfield{author}{\bibinfo{person}{Tong Li}, \bibinfo{person}{Zhaoyang Liu},
  \bibinfo{person}{Yanyan Shen}, \bibinfo{person}{Xue Wang},
  \bibinfo{person}{Haokun Chen}, {and} \bibinfo{person}{Sen Huang}.}
  \bibinfo{year}{2024}\natexlab{}.
\newblock \showarticletitle{Master: Market-guided stock transformer for stock
  price forecasting}. In \bibinfo{booktitle}{\emph{Proceedings of the AAAI
  Conference on Artificial Intelligence}}, Vol.~\bibinfo{volume}{38}.
  \bibinfo{pages}{162--170}.
\newblock


\bibitem[Liu(2024)]%
        {surveyfinancialaiarchitectures}
\bibfield{author}{\bibinfo{person}{Junhua Liu}.}
  \bibinfo{year}{2024}\natexlab{}.
\newblock \bibinfo{title}{A Survey of Financial AI: Architectures, Advances and
  Open Challenges}.
\newblock
\showeprint[arxiv]{2411.12747}~[q-fin.TR]
\urldef\tempurl%
\url{https://arxiv.org/abs/2411.12747}
\showURL{%
\tempurl}


\bibitem[Liu et~al\mbox{.}(2024)]%
        {liu2024itransformer}
\bibfield{author}{\bibinfo{person}{Yong Liu}, \bibinfo{person}{Tengge Hu},
  \bibinfo{person}{Haoran Zhang}, \bibinfo{person}{Haixu Wu},
  \bibinfo{person}{Shiyu Wang}, \bibinfo{person}{Lijun Ma}, {and}
  \bibinfo{person}{Mingsheng Long}.} \bibinfo{year}{2024}\natexlab{}.
\newblock \showarticletitle{iTransformer: Inverted Transformers Are Effective
  for Time Series Forecasting}.
\newblock \bibinfo{journal}{\emph{ICLR}} (\bibinfo{year}{2024}).
\newblock


\bibitem[Lo(2004)]%
        {lo2004adaptive}
\bibfield{author}{\bibinfo{person}{Andrew~W. Lo}.}
  \bibinfo{year}{2004}\natexlab{}.
\newblock \showarticletitle{The Adaptive Markets Hypothesis: Market Efficiency
  from an Evolutionary Perspective}.
\newblock \bibinfo{journal}{\emph{Journal of Portfolio Management}}
  (\bibinfo{year}{2004}).
\newblock


\bibitem[Lo et~al\mbox{.}(2000)]%
        {lo2000foundations}
\bibfield{author}{\bibinfo{person}{Andrew~W. Lo}, \bibinfo{person}{Harry
  Mamaysky}, {and} \bibinfo{person}{Jiang Wang}.}
  \bibinfo{year}{2000}\natexlab{}.
\newblock \showarticletitle{Foundations of Technical Analysis: Computational
  Algorithms, Statistical Inference, and Empirical Implementation}.
\newblock \bibinfo{journal}{\emph{The Journal of Finance}}
  (\bibinfo{year}{2000}).
\newblock


\bibitem[Mehrabian et~al\mbox{.}(2025)]%
        {mehrabian2025mamba}
\bibfield{author}{\bibinfo{person}{Ali Mehrabian}, \bibinfo{person}{Ehsan
  Hoseinzade}, \bibinfo{person}{Mahdi Mazloum}, {and} \bibinfo{person}{Xiaohong
  Chen}.} \bibinfo{year}{2025}\natexlab{}.
\newblock \showarticletitle{Mamba meets financial markets: A graph-mamba
  approach for stock price prediction}. In \bibinfo{booktitle}{\emph{ICASSP
  2025-2025 IEEE International Conference on Acoustics, Speech and Signal
  Processing (ICASSP)}}. IEEE, \bibinfo{pages}{1--5}.
\newblock


\bibitem[Nadaraya(1964)]%
        {nadaraya1964estimating}
\bibfield{author}{\bibinfo{person}{Elizbar~A Nadaraya}.}
  \bibinfo{year}{1964}\natexlab{}.
\newblock \showarticletitle{On estimating regression}.
\newblock \bibinfo{journal}{\emph{Theory of Probability \& Its Applications}}
  \bibinfo{volume}{9}, \bibinfo{number}{1} (\bibinfo{year}{1964}),
  \bibinfo{pages}{141--142}.
\newblock


\bibitem[Nie et~al\mbox{.}(2023)]%
        {Nie2023PatchTST}
\bibfield{author}{\bibinfo{person}{Yuqi Nie}, \bibinfo{person}{Nam H.~Nguyen},
  \bibinfo{person}{Phanwadee Sinthong}, {and} \bibinfo{person}{Jayant
  Kalagnanam}.} \bibinfo{year}{2023}\natexlab{}.
\newblock \showarticletitle{A Time Series is Worth 64 Words: Long-term
  Forecasting with Transformers}. In \bibinfo{booktitle}{\emph{International
  Conference on Learning Representations}}.
\newblock


\bibitem[Qin et~al\mbox{.}(2017)]%
        {qin2017alstm}
\bibfield{author}{\bibinfo{person}{Yao Qin}, \bibinfo{person}{Dongjin Song},
  \bibinfo{person}{Haifeng Cheng}, \bibinfo{person}{Wei Cheng},
  \bibinfo{person}{Guofei Jiang}, {and} \bibinfo{person}{Garrison~W.
  Cottrell}.} \bibinfo{year}{2017}\natexlab{}.
\newblock \showarticletitle{A dual-stage attention-based recurrent neural
  network for time series prediction}. In \bibinfo{booktitle}{\emph{Proceedings
  of the 26th International Joint Conference on Artificial Intelligence}}.
  \bibinfo{publisher}{AAAI Press}, \bibinfo{pages}{2627–2633}.
\newblock


\bibitem[Rossi(2013)]%
        {Rossi2013}
\bibfield{author}{\bibinfo{person}{Barbara Rossi}.}
  \bibinfo{year}{2013}\natexlab{}.
\newblock \showarticletitle{Exchange rate predictability}.
\newblock \bibinfo{journal}{\emph{Journal of economic literature}}
  \bibinfo{volume}{51}, \bibinfo{number}{4} (\bibinfo{year}{2013}),
  \bibinfo{pages}{1063--1119}.
\newblock


\bibitem[Vaswani et~al\mbox{.}(2017)]%
        {vaswani2017attention}
\bibfield{author}{\bibinfo{person}{Ashish Vaswani}, \bibinfo{person}{Noam
  Shazeer}, \bibinfo{person}{Niki Parmar}, \bibinfo{person}{Jakob Uszkoreit},
  \bibinfo{person}{Llion Jones}, \bibinfo{person}{Aidan~N Gomez},
  \bibinfo{person}{Lukasz Kaiser}, {and} \bibinfo{person}{Illia Polosukhin}.}
  \bibinfo{year}{2017}\natexlab{}.
\newblock \showarticletitle{Attention Is All You Need}.
\newblock \bibinfo{journal}{\emph{NeurIPS}} (\bibinfo{year}{2017}).
\newblock


\bibitem[Wu et~al\mbox{.}(2023)]%
        {wu2023timesnet}
\bibfield{author}{\bibinfo{person}{Haixu Wu}, \bibinfo{person}{Tengge Hu},
  \bibinfo{person}{Yong Liu}, \bibinfo{person}{Hang Zhou},
  \bibinfo{person}{Jianmin Wang}, {and} \bibinfo{person}{Mingsheng Long}.}
  \bibinfo{year}{2023}\natexlab{}.
\newblock \showarticletitle{TimesNet: Temporal 2D-Variation Modeling for
  General Time Series Analysis}. In \bibinfo{booktitle}{\emph{International
  Conference on Learning Representations}}.
\newblock


\bibitem[Wu et~al\mbox{.}(2021b)]%
        {wu2021autoformer}
\bibfield{author}{\bibinfo{person}{Haixu Wu}, \bibinfo{person}{Jiehui Xu},
  \bibinfo{person}{Jianmin Wang}, {and} \bibinfo{person}{Mingsheng Long}.}
  \bibinfo{year}{2021}\natexlab{b}.
\newblock \showarticletitle{Autoformer: Decomposition Transformers with
  Auto-Correlation for Long-Term Series Forecasting}. In
  \bibinfo{booktitle}{\emph{Advances in Neural Information Processing Systems
  (NeurIPS)}}.
\newblock


\bibitem[Wu et~al\mbox{.}(2021a)]%
        {wu2021tcts}
\bibfield{author}{\bibinfo{person}{Xueqing Wu}, \bibinfo{person}{Lewen Wang},
  \bibinfo{person}{Yingce Xia}, \bibinfo{person}{Weiqing Liu},
  \bibinfo{person}{Lijun Wu}, \bibinfo{person}{Shufang Xie},
  \bibinfo{person}{Tao Qin}, {and} \bibinfo{person}{Tie-Yan Liu}.}
  \bibinfo{year}{2021}\natexlab{a}.
\newblock \showarticletitle{Temporally correlated task scheduling for sequence
  learning}. In \bibinfo{booktitle}{\emph{Proceedings of the 38th International
  Conference on Machine Learning (ICML)}}.
\newblock


\bibitem[Yang et~al\mbox{.}(2020)]%
        {yang2020qlib}
\bibfield{author}{\bibinfo{person}{Xiao Yang}, \bibinfo{person}{Weiqing Liu},
  \bibinfo{person}{Dong Zhou}, \bibinfo{person}{Jiang Bian}, {and}
  \bibinfo{person}{Tie-Yan Liu}.} \bibinfo{year}{2020}\natexlab{}.
\newblock \showarticletitle{Qlib: An ai-oriented quantitative investment
  platform}.
\newblock \bibinfo{journal}{\emph{arXiv preprint arXiv:2009.11189}}
  (\bibinfo{year}{2020}).
\newblock


\bibitem[Ye and Keogh(2009)]%
        {ye2009}
\bibfield{author}{\bibinfo{person}{Lexiang Ye} {and} \bibinfo{person}{Eamonn
  Keogh}.} \bibinfo{year}{2009}\natexlab{}.
\newblock \showarticletitle{Time series shapelets: a new primitive for data
  mining}. In \bibinfo{booktitle}{\emph{Proceedings of the 15th ACM SIGKDD
  international conference on Knowledge discovery and data mining}}.
  \bibinfo{pages}{947--956}.
\newblock


\bibitem[Zeng et~al\mbox{.}(2023)]%
        {zeng2022dlinear}
\bibfield{author}{\bibinfo{person}{Ailing Zeng}, \bibinfo{person}{Muxi Chen},
  \bibinfo{person}{Lei Zhang}, {and} \bibinfo{person}{Qiang Xu}.}
  \bibinfo{year}{2023}\natexlab{}.
\newblock \showarticletitle{Are transformers effective for time series
  forecasting?}. In \bibinfo{booktitle}{\emph{Proceedings of the AAAI
  conference on artificial intelligence}}, Vol.~\bibinfo{volume}{37}.
  \bibinfo{pages}{11121--11128}.
\newblock


\bibitem[Zeng et~al\mbox{.}(2024)]%
        {ijcai2024p678}
\bibfield{author}{\bibinfo{person}{Liang Zeng}, \bibinfo{person}{Lei Wang},
  \bibinfo{person}{Hui Niu}, \bibinfo{person}{Ruchen Zhang},
  \bibinfo{person}{Ling Wang}, {and} \bibinfo{person}{Jian Li}.}
  \bibinfo{year}{2024}\natexlab{}.
\newblock \showarticletitle{Trade When Opportunity Comes: Price Movement
  Forecasting via Locality-Aware Attention and Iterative Refinement Labeling}.
  In \bibinfo{booktitle}{\emph{Proceedings of the 33rd International Joint
  Conference on Artificial Intelligence (IJCAI)}}. \bibinfo{pages}{6134--6142}.
\newblock
\href{https://doi.org/10.24963/ijcai.2024/678}{doi:\nolinkurl{10.24963/ijcai.2024/678}}
\newblock
\shownote{Main Track}.


\bibitem[Zhang et~al\mbox{.}(2020)]%
        {zhang2020doubleensemble}
\bibfield{author}{\bibinfo{person}{Chuheng Zhang}, \bibinfo{person}{Yuanqi Li},
  \bibinfo{person}{Xi Chen}, \bibinfo{person}{Yifei Jin},
  \bibinfo{person}{Pingzhong Tang}, {and} \bibinfo{person}{Jian Li}.}
  \bibinfo{year}{2020}\natexlab{}.
\newblock \showarticletitle{Doubleensemble: A new ensemble method based on
  sample reweighting and feature selection for financial data analysis}. In
  \bibinfo{booktitle}{\emph{Proceedings of the 2020 IEEE International
  Conference on Data Mining (ICDM)}}.
\newblock


\bibitem[Zhang et~al\mbox{.}(2017)]%
        {zhang2017stock}
\bibfield{author}{\bibinfo{person}{Liheng Zhang}, \bibinfo{person}{Charu
  Aggarwal}, {and} \bibinfo{person}{Guo-Jun Qi}.}
  \bibinfo{year}{2017}\natexlab{}.
\newblock \showarticletitle{Stock price prediction via discovering
  multi-frequency trading patterns}. In \bibinfo{booktitle}{\emph{Proceedings
  of the 23rd ACM SIGKDD international conference on knowledge discovery and
  data mining}}. \bibinfo{pages}{2141--2149}.
\newblock


\bibitem[Zhong et~al\mbox{.}(2023)]%
        {DTCC}
\bibfield{author}{\bibinfo{person}{Ying Zhong}, \bibinfo{person}{Dong Huang},
  {and} \bibinfo{person}{Chang-Dong Wang}.} \bibinfo{year}{2023}\natexlab{}.
\newblock \showarticletitle{Deep temporal contrastive clustering}.
\newblock \bibinfo{journal}{\emph{Neural Processing Letters}}
  \bibinfo{volume}{55}, \bibinfo{number}{6} (\bibinfo{year}{2023}),
  \bibinfo{pages}{7869--7885}.
\newblock


\bibitem[Zhou et~al\mbox{.}(2021)]%
        {zhou2021informer}
\bibfield{author}{\bibinfo{person}{Haoyi Zhou}, \bibinfo{person}{Shanghang
  Zhang}, \bibinfo{person}{Jieqi Peng}, \bibinfo{person}{Shuai Zhang},
  \bibinfo{person}{Jianxin Li}, \bibinfo{person}{Hui Xiong}, {and}
  \bibinfo{person}{Wenchao Zhang}.} \bibinfo{year}{2021}\natexlab{}.
\newblock \showarticletitle{Informer: Beyond Efficient Transformer for Long
  Sequence Time-Series Forecasting}.
\newblock \bibinfo{journal}{\emph{AAAI}} (\bibinfo{year}{2021}).
\newblock


\bibitem[Zhou et~al\mbox{.}(2022)]%
        {zhou2022fedformer}
\bibfield{author}{\bibinfo{person}{Tian Zhou}, \bibinfo{person}{Ziqing Ma},
  \bibinfo{person}{Qingsong Wen}, \bibinfo{person}{Xue Wang},
  \bibinfo{person}{Liang Sun}, {and} \bibinfo{person}{Rong Jin}.}
  \bibinfo{year}{2022}\natexlab{}.
\newblock \showarticletitle{Fedformer: Frequency enhanced decomposed
  transformer for long-term series forecasting}. In
  \bibinfo{booktitle}{\emph{International conference on machine learning}}.
  PMLR, \bibinfo{pages}{27268--27286}.
\newblock


\end{thebibliography}

\end{document}